\documentclass{article}
\usepackage{amsmath,graphicx}
\usepackage[preprint]{spconf}
\usepackage{verbatim}
\usepackage{ulem}
\usepackage{xcolor}
\usepackage[numbers]{natbib}
\setcitestyle{square}

\title{Themes Informed Audio-Visual Correspondence Learning}
\makeatletter
\def\@name{{Runze Su\textsuperscript{1, 2}}\sthanks{This work was done while the author was an intern at Kwai Y-tech lab.}, {Fei Tao\textsuperscript{1}} , {Xudong Liu\textsuperscript{3},Haoran Wei\textsuperscript{1, 4}},  {Xiaorong Mei\textsuperscript{3}, Zhiyao Duan\textsuperscript{1}},\\ {{Lei Yuan}\textsuperscript{1}}, {{Ji Liu}\textsuperscript{1}},  {{Yuying Xie}\textsuperscript{2}}\\}
\makeatother

\address{\textsuperscript{1}Seattle AI Lab, Kuaishou Technology\\\textsuperscript{2}Department of Statistics and Probability \& \\
Department of Computational Mathematics, Science and Engineering, Michigan State University\\ \textsuperscript{3}Ads Platform, Kuaishou Technology\\ 
\textsuperscript{4}Department of ECE, University of Texas at Dallas}

\begin{document}
\maketitle
\copyrightnotice{\copyright\ IEEE 2021}
\toappear{To appear in {\it Proc.\ ICASSP 2021, 6-11 June 2021, Toronto, Ontario, Canada}}
\begin{abstract}
The applications of short-term user-generated video (UGV), such as Snapchat, and Youtube short-term videos, booms recently, raising lots of multimodal machine learning tasks. Among them, learning the correspondence between audio and visual information from videos is a challenging one. Most previous work of the audio-visual correspondence(AVC) learning only investigated constrained videos or simple settings, which may not fit the application of UGV. In this paper, we proposed new principles for AVC and introduced a new framework to set sight of videos' themes to facilitate AVC learning. We also released the KWAI-AD-AudVis corpus which contained 85432 short advertisement videos (around 913 hours) made by users. We evaluated our proposed approach on this corpus, and it was able to outperform the baseline by 23.15\% absolute difference.
\end{abstract}
\vspace{-4 mm}

\section{Introduction}
\vspace{-3 mm}
\label{sec:intro}
Recently, the applications of short-term user-generated video (UGV) booms fast, such as Youtube short-term videos, Snapchat and Kwai. Audio-visual correspondence (AVC) learning, which can tell whether or how well the audio and visual information matches in the video, can bring benefits to these applications. It can recommend audio or visual streams to users given the other modality for being more ``like", evaluate the quality of the short-term video for pushing to users, and build a better high-level representation of videos for other uses. 

Efforts have been made on the AVC learning. However, most previous works have limitations, mainly lying on two shortages: the task setting was simple, such as matching background audio to single image \cite{aytar2016soundnet}; and the approaches relied on simple assumptions that audio and visual information should be similar in some projected space \cite{arandjelovic2017look,li2019query}. These 
shortcomings may fail the systems on UGVs, whose information is complex and complicated. One example can illustrate this: the user may combine a series of cheerful wedding ceremony photos and low-spirited style music to present the theme “marriage is the tomb of love” -- spoken by Giacomo Casanova.

To model the complex relationship between visual and audio information, we introduce the concept of theme and propose a Theme
Informed AVC (Ti-AVC) learning algorithm. This idea is based on the following thoughts. All  UGVs convey the authors' ideas (video themes), which may consist of several aspects. Each aspect is reflected by one or more modalities. Since the theme can contain complex semantic information, the aspects may disagree with each other. Simply measuring agreement or similarity of the modalities cannot tell how they match. The matched audio and visual information should follow two principles: 1) they need to convey the same desired theme together; 2) there should be a relationship between them when present the theme. For the first principle, we designed a a novel framework to inject the theme information into AVC learning. Since it is not clear how to represent the theme, we adopted the video tags to direct model the theme indirectly in this paper. For the second principle, we followed conventional ideas and adopted a state-of-the-art framework to model the relationship.

To evaluate our proposed framework, we collected 85,432 UGVs from Kwai, a popular short-term video app in China. All the collected videos are advertisement (ads) uploaded by commercial advertisers. We will publish the dataset as the extension of KWAI-AD \cite{chen2020imram} dataset. In this dataset, our proposed approach gained 23.15\% improvement in accuracy AUC compared to a state-of-the-art AVC learning framework.

We summarize the contribution of this paper below: 1) We introduced new principles for the AVC learning task. 2) We proposed the first theme informed audio-visual correspondence (Ti-AVC) framework which is suitable for UGVs. It outperformed the state-of-the-art baseline by 23.15\% absolute difference, and its hidden values indicate the modality information flow in AVC. 3) We published the first audio-visual dataset grouped by contents based on short-term ads video. 

\vspace{-4 mm}
\section{Related Work}
\vspace{-3 mm}
\label{sec:related}
Researchers cast much attention on reciprocity between audio and visual information on various tasks \cite{tao2020end}. Although transfer learning has been proposed to convey information across modalities \cite{zhao2018sound, Tao_2018_2}, how to model the correspondence between modalities is still an open question.

$L^3$ net \cite{aytar2017see} was proposed to explicitly model AVC. It used several sub-networks to perform inputs processing and modalities fusion. Relying on the max-pooling layer in the fusion sub-network, the $L^3$ net had a flexible framework that was able to take sequential or single input. It showed state-of-the-art performance and a new perspective to perform sound localization task\cite{zhao2018sound, arandjelovic2018objects, wu2019dual}. In audio-visual cross-modal embedding designs, the pre-trained $L^3$ net is deployed as embedding extractor \cite{cramer2019look, chung2019perfect}. Verma et al. applied $L^3$ net framework to learn AVC based on the emotion from audio and visual streams\cite{verma2019learning}. This work was evaluated on a new released dataset that contained audio and visual emotion information. To increase the correspondence, gating mechanism \cite{Tao_2018_4} was applied to filter uncorresponding information in audio and visual streams; dual attention matching \cite{wu2019dual} added attention to both audio and visual inputs to predict their event sequential localization relevance between modals; elastic multi-way network \cite{wang2019novel} designed a loss function with the distance between samples and an anchor point to encourage correspondence; \cite{Tao_2018_6} relied on a bimodal recurrent neural network to learn the temporal correspondence information in a data-driven fashion. Unsupervised methods such as video audio correspondence were also investigated such as audio-visual deep clustering model\cite{lu2019audio}. Most of the approaches focused on modeling the similarity between modalities and showed decent performance. However, most of the approaches were only evaluated on constrained dataset. AVC learning on unconstrained data is still a complicated and difficult task. \cite{zhu2020deep, baltruvsaitis2018multimodal}.

There are several public available unconstrained audio-visual datasets such as UGVs datasets, but none of them are suitable for the short-term videos case. Specifically, Youtube-8M \cite{abu2016youtube} only, one of the most popular UGVs dataset, covers various themes (i.e. tags), but its video quality is not controlled intentionally. Also, the video duration in Youtube-8M exceeds the typical length for a short-term video. On the other side, the Moments in Time \cite{monfort2019moments} contains 1,000,000 3-second videos, which is too short. Flickr-SoundNet\cite{aytar2016soundnet} is a unconstrained dataset, however it only has single image with background sound track. The shortage of good quality short-term UGVs inspired us to collect a new data, whose details will be introduced later.

\vspace{-4 mm}
\section{KWAI-AD-AudVis Dataset}
\vspace{-3 mm}
\label{sec:data}
In this study, we developed our framework on KWAI-AD-AudVis dataset. It constists of 85,432 ads videos (around 913 hours) from the China popular short-term video app, Kwai. The videos were made and uploaded by commercial advertisers rather than personal users. The reason to use the ads videos lied on two folds: 1) the source guarantees videos under control to some level, such as high-resolution pictures and intentionally designed scene; 2) ad videos simulate audio-visual matching style as manually composited by users in Kwai app. It can be seen as a quality controlled UGVs dataset.

In the KWAI-AD-AudVis dataset, each UGV/ad has a label for the industry category. The videos were randomly picked from a pool. The pool was formed by selecting the ads from several contiguous days. We estimate the number of clicks advertisers receive every time the ads come out as a criteria during collection. In this dataset, half of the ads have a high rate to raise customers' interests in the products, and the other half has a relatively low attraction. The short videos have been classified into 19 themes by uploaders with an average length of seconds.  The audio track had 2 channels (we mixed to mono channel in the study) and was sampled at 44.1 kHz, while the visual track had a resolution of $720 \times 1280$ and was sampled at 25 frames per second (FPS). This dataset is an extension of the KWAI-AD corpus \cite{chen2020imram}. It is not only suitable for tasks in the multimodal learning area, but also for ones in ads recommendation.

The details and data of KWAI-AD-AudVid can be accessed through Zenodo\footnote{https://zenodo.org/record/4023390\#.X12Dr5NKgUE}. It shows that the ads videos have three main characteristics: 1) The videos may have very inconsistent information in visual or audio streams. For example, the video may play a drama-like story at first, and then present the product introduction, whose scenes are very different. 2) The correspondence between audio and visual streams is not clear. For instance, similar visual objects (e.g. talking salesman) come with very different audio streams. 3) The relationship between audio and video varies in different industries. For example, games or E-commerce ads will have very different styles. These characteristics make the dataset suitable yet challenging for our study on AVC learning.

 

\vspace{-4 mm}
\section{Proposed Approaches}
\vspace{-3 mm}
\label{sec:approaches}

\subsection{Data and Feature}
\vspace{-2 mm}
In this study, we used KWAI-AD-AudVis dataset to develop our AVC learning framework. To reduce the data size and training workload, we used our in-house key-frame extractor to extract 8 frames from each video to represent the visual information. The resulting images did not follow the original order in the video. Audio tracks were extracted as same as in original videos. The visual and audio information are pre-processed through Mobilenetv2 \cite{sandler2018mobilenetv2} and VGGish \cite{hershey2017cnn}. Embedding from top layers of the two pre-trained was fed to our proposed system.

\vspace{-2 mm}
\subsection{Theme Informed AVC learning System}
\begin{figure}
    \centering
    \includegraphics[scale=0.4]{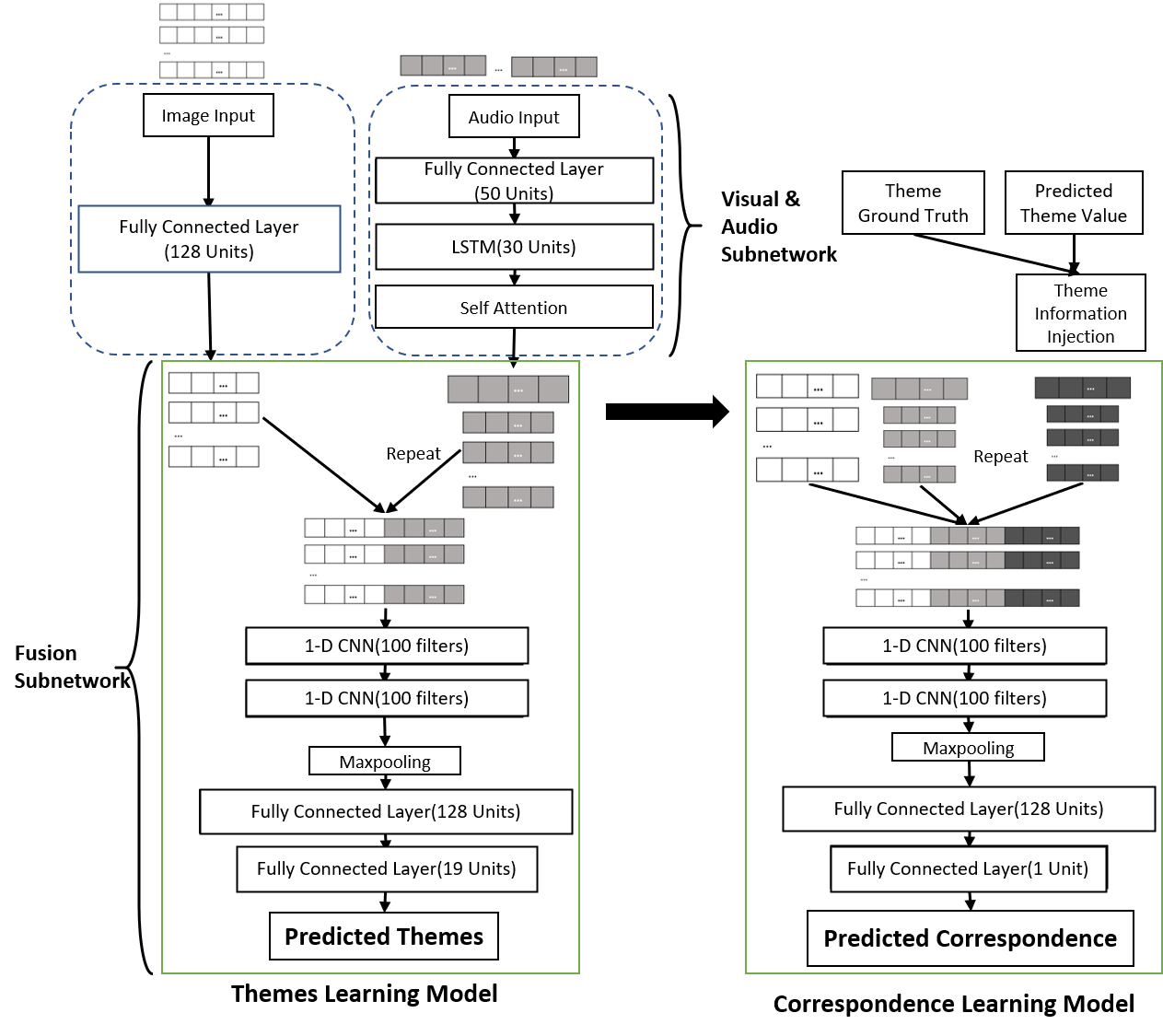}
    \caption{Diagram of the proposed framework.}
    \label{fig:maker1}
\end{figure}

\vspace{-2 mm}
Figure \ref{fig:maker1} shows the diagram of our proposed approach, theme-informed audio-visual correspondence (Ti-AVC) learning framework. It consisted of two parts, a theme-learning (TL) model and a correspondence-learningCL model.

For the TL model, we were inspired by $L^3$ net and designed a similar network as $L^3$ net, except its task was theme prediction (in this study, it is ads industry category prediction). It took audio and visual embedding as input. It consisted of three sub-networks. Two sub-network processed the input of single modality separately, and the third one processed the fused information. The audio sub-network had a time distributed dense layer, an LSTM layer and a self-attention layer. Its output is a 128-D vector. The visual sub-network had a fully connected layer, whose parameters were shared across different input frames. Its output was a sequence of 8 128-D vectors. The output from the audio sub-network was repeated and concatenated to each vector from the visual sub-network. The concatenated embedding was fed into the fusion sub-network. The fusion sub-network had two 1-D convolutional neural networks (CNN), a max-pooling layer and 2 fully connected (FC) layers to predict themes.


Once the TL model was trained, we fixed it as an embedding extractor to extract three types of information for the CL model: audio embedding from the top-layer of the audio sub-network, visual embedding from the top-layer of the visual sub-network and the predicted theme. We concatenated the predicted theme with the theme ground-truth to form the theme information, which was injected into the CL model with audio and visual embedding. By adding the true theme, we expect the CL model to learn how the input audio and visual embedding performs in predicting the theme. This was following the principle (1) we mentioned in section \ref{sec:intro}. In this study, the theme ground-truth corresponded to the visual modality. The CL model has similar architecture as the fusion sub-network in the TL model. We intended to use both of the theme prediction and ground-truth to tell how the two modalities represent the desired theme. The CL model was expected to capture two points: 1) how the desired theme was presented; 2) how the modalities related to each other. These two points corresponded to the two principles we proposed in Section \ref{sec:intro}. The correspondence result was eventually predicted based on the two points. We would like to point out that it was not cheating to have the theme ground-truth in the CL model, since we assumed this information available during AVC learning.

\vspace{-4 mm}
\section{Experiment and Analysis}
\vspace{-3 mm}
\label{sec:experiment}

\subsection{Experiment Setup}
\vspace{-2 mm}
We used the original videos from the KWAI-AD-AudVis dataset as positive samples, where we assumed audio and visual information matched with each other. Negative samples were generated by pairing audio and visual tracks from different videos. We generated the same number of negative samples as positive ones. The dataset was partitioned to 80\%, 10\% and 10\% for training, validation and testing respectively. We applied Adam as the optimizer and set the learning rate as 0.0001, batch size as 8 for both models in all experiments. We also set an early stopper, which monitors loss on the validation set. If validation loss doesn't decrease for consecutive 5 epochs, the training will stop. We used ads industries categories as theme information in this study.

We built two baselines for comparison. The first baseline (denoted as ``baseline-1") borrows the architecture from the themes learning model. To make a fair comparison, we made two adjustments for correspondence learning: 1) we replaced the theme prediction task by correspondence prediction; 2) we doubled the number of all trainable layers in the fusion sub-network to guarantee the same parameter size as our proposed approach. The second baseline (denoted as ``baseline-2") had exactly the same architecture as baseline-1, except we input theme ground-truth to the fusion sub-network concatenated with the modalities embedding. This made the comparison fair since the system also got theme information like the proposed approach. For the proposed approach, we made two training strategies and therefore had two systems. We named the system that trained TL and CL models separately as ``Ti-AVC", while we named the one jointly trained (i.e. a multitask learning system with TL and CL tasks) as``joint Ti-AVC". We kept all systems having the same number of parameters.

\vspace{-2 mm}
\subsection{Experiment Results}
\vspace{-2 mm}
The accuracy AUC score of each system is shown in Table \ref{tab:results}. The baseline-1, which had similar architecture to $L^3$ net, had random-guess results (we had the same amount of positive and negative samples). The baseline-2, which was the same as baseline-1 except it took the theme as an extra input, could outperform baseline-1 by 18.94\%. This verified our hypothesis that theme information was necessary for AVC learning on UGVs. Both of our proposed approaches beat the baselines (by at least 3.36\% absolute difference) with the Ti-AVC achieving the best performance. Since the TL and CL models were trained separately in Ti-AVC, it indicates that properly injecting the information on how the audio and visual modalities presented the desired theme could improve the performance of the correspondence learning task. We would like to emphasize that the Ti-AVC is flexible in application. The TL model can be fixed as embedding extractor and the theme categories provided by CL model can be obtained from either modality (in this study, we made it follow visual modality).

\begin{figure}[tb]
    \centering
    \hspace*{-5mm}\includegraphics[width=1.0\columnwidth, height=6.5cm]{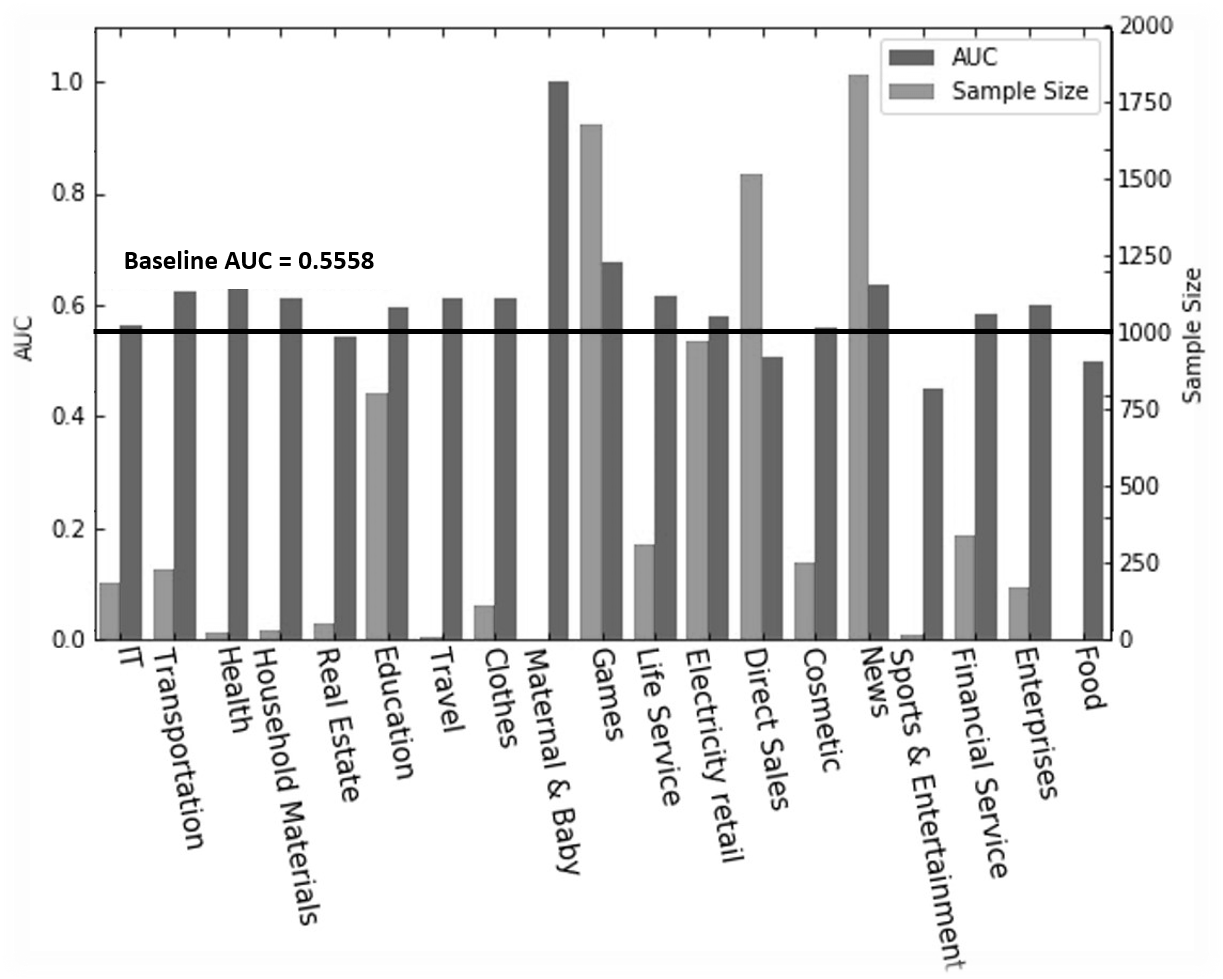}
    \caption{AUC and sample counts per ADs category. The dark grey bar represents the AUC, whose scale axis is on left; the light grey bar represents the number of samples, whose scale axis is on right. The horizontal line is the baseline-1 accuracy AUC.}
    \label{fig:cateauc}
\end{figure}

We also performed evaluations within each theme category (shown in Figure \ref{fig:cateauc}), where all the testing candidates were from the same category in AUC computation. Since all the testing candidates had the same theme ground-truth, the evaluation was equivalent to eliminating the information of theme ground-truth. The CL model can only obtain help from the difference between the theme prediction and ground-truth. This difference can represent "how the desired themes are presented" (principle (1) in Section \ref{sec:intro}), so the results can reflect the effectiveness of the principle (1) in the AVC task. We compare the results with the baseline-1 (the horizontal line in Figure \ref{fig:cateauc}), which did not include theme ground-truth during inference. The result shows that the Ti-AVC framework dominates the baseline in 15 categories out of 19. Especially, we notice all the categories with most samples outperformed the baseline. This result indicates that the proposed framework can help improve the correspondence learning even without theme information, and justify the proposed principle (1).

\begin{table}[tb]
    \centering
    \begin{tabular}{c|c}
        \hline
        Model & Match AUC\\
        \hline
        Baseline-1 & 55.58\%\\
        Baseline-2 & 74.52\%\\
        \hline
        Joint Ti-AVC & 77.88\%\\
        Ti-AVC  & \textbf{78.73\%}\\
        \hline
    \end{tabular}
    \caption{Summary of experiment results.}
    \label{tab:results}
\end{table}

\vspace{-2 mm}
\subsection{Contribution Analysis}
\vspace{-2 mm}

To further verify the rationality of our proposed approach, we analyzed the contribution of each input in the CL model. We use the information flow fed into the convolutional layers to indicate the importance of each modal. As both positive and negative values have an effective influence on the prediction, we set the absolute value of the input values as our estimation statistic. Define the contribution in equation \ref{eq:contrib}, where $W_{i}$ is the weight of the first layer connecting the $i_{th}$ input and $X_{i}$ is the $i_{th}$ input, $I$ is the input type (audio, visual, predicted theme and true theme).
\begin{equation}
    \label{eq:contrib}
    Contribution^{I} = \sum|{W_{i}^{I}\cdot X_{i}^{I}}|
\end{equation}

Table \ref{tab:table3} listed the computed proportion of the inputs for the matched pairs. It showed that audio modalities have the most portion contribution (58.78\%). The theme information took up 10.38\%, where the predicted and true ones were close (4.52\% and 5.86\%). This result indicated both of them could not be neglected, which verified our proposed principles for AVC in section \ref{sec:intro} and the capability of the proposed approach. 

\begin{table}
    \centering
    \begin{tabular}{|c|c|c|c|}
        \hline
         Vision  & Audio& Predicted Themes  & True Themes \\
        \hline
        30.85\% & 58.78\% & 4.52\% & 5.86\% \\
        \hline
    \end{tabular}
    \caption{Modal contributions calculated from a batch of positive audio-visual pairs and a batch of negative audio-visual pairs. }
    \label{tab:table3}
\end{table}

\vspace{-3 mm}
\section{Conclusion}
\vspace{-3 mm}
\label{sec:conclusion}

In this paper, we proposed new principles in audio-visual correspondence learning on users generated videos, which introduced theme information in AVC tasks. We proposed a new framework to perform the AVC task under unconstrained scenarios. To evaluate the proposed approach, we also collected and released the KWAI-AD-AudVis corpus, consisting of 85432 short-term videos (around 913 hours). 

Our proposed approach was able to outperform a state-of-the-art AVC framework by 23.15\% in accuracy AUC. We also showed that the proposed approach could still outperform the baseline even without theme information. Besides, the proposed framework would be flexible in real application as the TL model can be fixed and the theme information can correspond to either modality. This study only focused on learning correspondence between audio and visual modalities by concatenating the embedding of the modalities. The future work lies on a more sophisticated fusion strategy and further analyzing how the modality correlates with each other.

\vfill\pagebreak
\bibliographystyle{IEEEtran}
\bibliography{refs}

\end{document}